\def\ARXIV{}
\documentclass[runningheads]{llncs}
\usepackage[T1]{fontenc}
\usepackage{graphicx}
\usepackage{booktabs}
\usepackage{amsmath}
\usepackage{amssymb}
\usepackage{multirow}
\usepackage{xcolor}
\usepackage{url}
\usepackage[hidelinks,hypertexnames=false]{hyperref}  
\usepackage{tikz}
\usetikzlibrary{arrows.meta,positioning,calc,fit,backgrounds}
\usepackage{float}     
\DeclareRobustCommand{\Letter}{\tikz[baseline=-0.15ex]{\draw[line width=0.45pt] (0,0) rectangle (0.72em,0.5em);\draw[line width=0.45pt] (0.02em,0.48em) -- (0.36em,0.2em) -- (0.7em,0.48em);}}

\graphicspath{{figures/}}

\newcommand{\ppos}{P(\mathrm{pos})}

\begin{document}

\title{The Score Granularity Gap in Black-Box\\ LLM Classification: A Comparative Study\\ of Confidence Constructions}
\titlerunning{The Score Granularity Gap}
\author{Ao Sun\texorpdfstring{\textsuperscript{(\Letter)}}{} \and Tian Sun \and Jiaxing Geng}
\authorrunning{A. Sun et al.}
\institute{Independent Researchers\\
\email{ao.sun@outlook.com}, \email{honeytiansun@gmail.com}, \email{gengjiaxing95@gmail.com}}

\maketitle

\begin{abstract}
Large language models (LLMs) are increasingly deployed as black-box classifiers in pipelines that automate confident decisions and route uncertain ones to human review. Such \emph{selective prediction} needs a confidence score that an operator can threshold at a chosen risk level. Prior work asks whether LLM confidence is well calibrated or well ranked; we ask a complementary, deployment-oriented question that has been largely overlooked: at what \emph{resolution} can the score be thresholded? We call the answer the \emph{score granularity gap}. Through a controlled comparison of seven ways to build a confidence score---from a single verbalized number, to token probabilities, to querying the model many times and combining the answers---across 25 model--dataset pairs (9 LLMs, 3 benchmarks), we find that single-shot verbalized confidence, once correctly converted to a class probability, ranks cases surprisingly well, yet collapses to only a handful of acceptance levels. It therefore offers an operator only a few coarse thresholds, no matter how well it ranks. We show which constructions widen this gap, at what inference cost, and with what effect on ranking---notably that multi-query aggregation helps weak models but can \emph{degrade} already-strong ones. We translate these trade-offs into concrete deployment guidance.

\end{abstract}

\section{Introduction}
\label{sec:intro}

Large language models are increasingly used as black-box classifiers across high-stakes domains, from medical triage to content moderation. When such a system runs without per-call access to model internals, practical deployment hinges on one capability: the model must attach to each prediction a \emph{confidence score} that an operator can compare against a threshold, automating the cases above the threshold and escalating the rest to a human reviewer. This is the classical \emph{selective prediction} setting, and its central control is the threshold: moving it trades \emph{coverage} (the fraction of cases handled automatically) against \emph{risk} (the error rate among them).

A long line of work asks whether LLM confidence is trustworthy in this role---whether it is \emph{calibrated} (does a stated $80\%$ mean $80\%$ correct?) and whether it \emph{ranks} cases correctly---and recurrently finds verbalized confidence overconfident. We argue that a prior, largely overlooked question matters just as much: not \emph{whether} the score is reliable, but at what \emph{resolution} it can be acted upon. A score may order cases almost perfectly yet be useless for fine-grained control if it takes only a few distinct values---an operator who wants the most confident $70\%$ has no threshold that selects exactly $70\%$. The reachable operating points form a coarse staircase, not a smooth dial.

We call this the \textbf{score granularity gap}, and a deployment-oriented study of LLM confidence must measure it directly. Doing so requires looking at a confidence score along three distinct axes, which we state here in plain terms and formalize in Section~\ref{sec:setup}: (i)~\emph{ranking quality}---does the score order cases well, so that a threshold separates good decisions from bad ones? (ii)~\emph{score resolution}---how many distinct, well-spread thresholds does the score actually offer? and (iii)~\emph{cost and interpretability}---how many model calls does the score cost, and can a human read why it was assigned? A score is deployment-ready only when it scores well on the axes the deployment cares about; no single axis suffices.

Against these axes we compare the natural ways to build such a score---asking the model directly, reading its token probabilities, or querying it several times (with paraphrases or interpretable sub-questions) and combining the answers. Our central finding: direct verbalized confidence, once converted to a class probability, ranks cases surprisingly well but is far too coarse for threshold control; token probabilities with calibration widen the resolution cheaply where available; and multi-query aggregation widens it most, at several calls per input and---counter-intuitively---sometimes \emph{degrading} already-strong models.

\smallskip
\noindent\textbf{Contributions.}
\begin{enumerate}
\item We identify and operationalize the \textbf{score granularity gap}---why a well-ranking LLM confidence score can still be unusable for threshold-based deployment---with metrics for both how many distinct values a score takes and how they are distributed (Section~\ref{sec:setup}).
\item We present a \textbf{controlled comparison of seven confidence constructions} across 25 model--dataset pairs, isolating each ingredient---conversion, token probabilities, calibration, multiple queries, and learned vs.\ naive combination (Sections~\ref{sec:method}--\ref{sec:experiments}).
\item We show that the popular ``LLM confidence is unreliable'' narrative is partly an artifact of measurement (conflating confidence-in-the-answer with class probability) and of resolution, and that \textbf{multi-query aggregation helps weak models but can hurt strong ones}---a trade-off we make explicit rather than averaging away (Section~\ref{sec:experiments}).
\item We give \textbf{deployment guidance}, including an inference-cost accounting and the case for interpretable sub-task scores as a deployment property, not a ranking advantage (Sections~\ref{sec:interp} and~\ref{sec:reco}).
\end{enumerate}

\section{Related Work}
\label{sec:related}

\paragraph{Uncertainty elicitation.} Verbalized confidence is often overconfident~\cite{xiong_can_2024}; self-evaluation~\cite{kadavath_language_2022} and semantic uncertainty~\cite{kuhn_semantic_2023} offer alternatives, and models can be trained to express calibrated uncertainty~\cite{lin_teaching_2022}. Recent studies examine the gap between token-level and verbalized signals~\cite{zhou_relying_2024}, benchmark verbalized reliability~\cite{yang_verbalized_nodate}, and estimate confidence under black-box access~\cite{pedapati_large_2025}. These works ask \emph{whether} LLMs can be confident; we ask \emph{at what resolution} a deployable threshold can be placed.

\paragraph{Calibration.} Temperature scaling~\cite{guo_calibration_2017} and beta calibration~\cite{kull_beta_2017} adjust probability estimates; few-shot LLMs in particular need calibration~\cite{zhao_calibrate_nodate}, and a recent survey maps the area~\cite{wang_calibration_2023}. Calibration is orthogonal to our concern: a monotone recalibration cannot create distinct values, so a calibrated score taking four values still offers four thresholds.

\paragraph{Selective prediction and decomposition.} Selective classification trades coverage for risk~\cite{geifman_selective_2017}; conformal methods add distribution-free guarantees~\cite{angelopoulos_conformal_2025}; and recent work surveys LLM abstention~\cite{wen_know_nodate}. Decomposed prompting~\cite{khot_decomposed_2023} and chain-of-thought~\cite{wei_chain--thought_2023} use sub-steps to reach a \emph{better answer}; we instead repurpose decomposition to \emph{build a score}, with sub-tasks as features rather than reasoning steps. Estimating confidence from agreement across rephrased queries~\cite{portillo_wightman_strength_2023,wang_multi-perspective_2024} is closest to our multi-prompt construction; we additionally study learned combination, interpretable sub-tasks, and the resolution axis. Extracting structural signals for single-round confidence~\cite{yang_trust_2026} is related to our sub-task variant.

\section{Problem Setup}
\label{sec:setup}

We scope the study to \emph{binary} classification, the simplest setting in which the granularity gap appears: a single scalar score suffices and a threshold reduces to one operating point. Constraining every signal to the shared interval $[0,1]$ also makes learned combination and threshold selection well defined; open-ended generation provides no such common coordinate system. We return to this choice in Section~\ref{sec:limits}.

\subsection{Confidence and Its Conversion to a Class Probability}
\label{sec:conf}

For an input $x$, a black-box LLM returns a predicted label $\hat{y}\in\{0,1\}$ and a \emph{verbalized confidence} $c\in[0,1]$. Crucially, $c$ is the model's confidence \emph{in the label it chose}, not the probability of the positive class. The two differ whenever the model predicts the negative class: a confident ``no'' ($\hat y{=}0,\,c{=}0.95$) is a \emph{low} class probability, while a hesitant ``yes'' ($\hat y{=}1,\,c{=}0.55$) is a moderate one. Class-ranking and calibration metrics operate on the class probability, and selective prediction thresholds the confidence derived from it (Section~\ref{sec:deploy}), so we first convert
\begin{equation}
    s \;=\; \ppos \;=\;
    \begin{cases} c & \text{if } \hat{y} = 1,\\[2pt] 1 - c & \text{if } \hat{y} = 0,\end{cases}
    \label{eq:conversion}
\end{equation}
and call the converted single-shot score \textbf{Verb}. The class probability $s$ is the common currency of what follows: classification metrics and calibration act on $s$, while the acceptance rule of Section~\ref{sec:deploy} acts on the prediction confidence derived from it. In our data $c\ge 0.5$ for over $97\%$ of examples. Prior work often evaluates confidence in the chosen answer~\cite{xiong_can_2024,tian_just_2023}---appropriate for answer-level uncertainty, but overstating ``unreliability'' when a class probability is what the task needs.

\subsection{Thresholded Deployment: Operating Points and Deployment Properties}
\label{sec:deploy}

Given a score $s$, selective prediction uses the thresholded label $\hat y_s=\mathbf{1}[s\ge 0.5]$ and the \emph{confidence in that prediction}, $q=\max(s,1-s)$. Given a threshold $\tau$, the system \emph{accepts} (predicts automatically) every case with $q\ge\tau$ and \emph{abstains} on the rest---a two-sided rule under which a confident negative is accepted exactly like a confident positive. Risk--coverage quantities score correctness against $\hat y_s$ for every construction; for Verb this agrees with the model's own label $\hat y$ whenever $c\ge0.5$ (${>}97\%$ of cases). Reported accuracies are those of $\hat y$. Sweeping $\tau$ traces the \textbf{risk--coverage curve}: each $\tau$ yields a \emph{coverage} (fraction accepted) and a \emph{risk} (error rate among accepted). An \textbf{operating point} is one achievable (coverage, risk) pair. Operators usually fix an \emph{error budget}---a maximum tolerable risk, e.g.\ $5\%$---and want the highest coverage achievable within it; we report this as $\mathrm{cov}@\tau$ (coverage at error rate $\le\tau$). We summarize the whole curve by the \textbf{AURC} (area under the risk--coverage curve; lower is better) and the ranking itself by \textbf{PR-AUC}.

We use the term \textbf{deployment property} for any attribute of a confidence score, \emph{beyond} expected accuracy, that determines whether it can be operated in practice. Three such properties drive this paper: \emph{ranking quality} (Section~\ref{sec:resolution}), \emph{score resolution} (next), and \emph{interpretability}---whether a human can read why a given score was assigned, which is a requirement, not a luxury, in regulated domains (Section~\ref{sec:interp}).

\subsection{Measuring Score Resolution: Count, Spread, and the Operating-Point Bound}
\label{sec:resolution}

A score that ranks well is not yet thresholdable. Since the acceptance rule of Section~\ref{sec:deploy} thresholds the prediction confidence $q=\max(s,1-s)$, resolution must be measured on $q$, \emph{not} on $s$: two class probabilities such as $s{=}0.1$ and $s{=}0.9$ are distinct values of $s$ but collapse to the single acceptance level $q{=}0.9$. Under deterministic thresholding (no randomized tie-breaking), every test case sharing a $q$ value must be accepted or abstained together, so a score whose $q$ takes $V$ distinct values induces at most $V$ non-trivial operating points---and a deployment that can only set thresholds from a finite menu of size $K$ reaches at most $\min(V,K)$ of them. \textbf{The score-side bound $V$ caps controllability independently of how well the score ranks.} This is the crux of the granularity gap: a coarse score is \emph{score-limited} however fine the menu; a fine score is limited only by the menu.

We measure resolution with two complementary quantities, both computed on the test-set values of $q$:
\begin{itemize}
\item \textbf{Granularity} $G_q=\bigl|\{q(x):x\in\mathcal{D}_{\text{test}}\}\bigr|$: the number of distinct acceptance levels---the score-side bound $V$ above. We count distinct values after rounding to four decimals, the finest resolution at which an operator would plausibly place a threshold, so floating-point jitter does not count as resolution.
\item \textbf{Score entropy} $H_q=-\sum_{b=1}^{20} p_b \log p_b$: bin the $q$ values into $20$ equal-width bins over $[0.5,1]$, let $p_b$ be the fraction of test cases in bin $b$, and take the Shannon entropy of this histogram, in \emph{nats} ($0$ means all mass in one bin, $\log 20\approx 3.0$ means perfectly uniform). $H_q$ captures \emph{how evenly} the levels are used, which the count alone does not, and guards against inflating $G_q$ with values packed into a narrow band.
\end{itemize}
Both, like any histogram statistic, are properties of the score's parameterization at the stated precision, not invariants of its ranking.

\section{Confidence Constructions}
\label{sec:method}

\begin{figure}[tbp]
\centering
\resizebox{\textwidth}{!}{%
\begin{tikzpicture}[
    font=\small,
    box/.style={draw, rounded corners, align=center, minimum height=8mm, inner sep=3pt},
    llm/.style={box, fill=blue!8},
    op/.style={box, fill=orange!12},
    res/.style={box, fill=green!10},
    >=Stealth, node distance=6mm and 9mm,
]
\node (x) [box, fill=gray!8] {Input $x$};

\node (llm1) [llm, above right=4mm and 10mm of x] {LLM, 1 call};
\node (conv) [op, right=of llm1] {convert\\Eq.~\eqref{eq:conversion}};
\node (verb) [res, right=of conv] {\textbf{Verb} $s$\\(or \textbf{logprob})};

\node (llmM) [llm, below right=4mm and 10mm of x] {LLM, $m$ calls\\(sub-tasks \emph{or}\\paraphrases)};
\node (vec)  [op, right=of llmM] {feature vec.\\$\mathbf{x}{=}(x_1,\dots,x_m)$};
\node (agg)  [op, right=of vec] {aggregate\\(avg \emph{or} LR)};
\node (cal)  [op, right=of agg] {temp.\\scale};
\node (subc) [res, right=of cal] {\textbf{Sub-cal}/\\\textbf{MP-LR} $s$};

\draw[->] (x) |- (llm1);
\draw[->] (llm1) -- (conv);
\draw[->] (conv) -- (verb);
\draw[->] (x) |- (llmM);
\draw[->] (llmM) -- (vec);
\draw[->] (vec) -- (agg);
\draw[->] (agg) -- (cal);
\draw[->] (cal) -- (subc);

\node (thr) [box, fill=red!8, right=14mm of $(verb)!0.5!(subc)$] {threshold $\tau$ on\\$q{=}\max(s,1{-}s)$:\\accept / abstain};
\draw[->] (verb) -| (thr);
\draw[->] (subc) -| (thr);
\end{tikzpicture}%
}
\caption{The confidence constructions we compare. \textbf{Single-shot} (top, $1$ call): the model's verbalized confidence is converted to a class probability (\textbf{Verb}), or first-token Yes/No log-probabilities are used (\textbf{logprob}). \textbf{Multi-signal} (bottom, $m{=}10$ calls): the model answers $m$ interpretable sub-tasks \emph{or} $m$ paraphrases of the task; the $m$ scalar scores form a feature vector that is aggregated (simple average or learned logistic regression, LR) and optionally temperature-scaled. Any resulting score $s$ yields a prediction confidence $q=\max(s,1-s)$, which is thresholded for selective prediction.}
\label{fig:pipeline}
\end{figure}
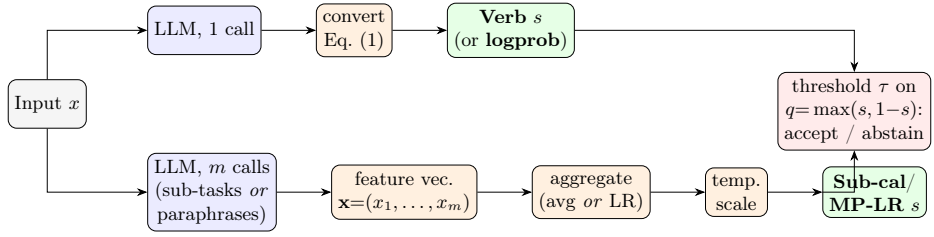

We compare seven ways to turn black-box LLM outputs into a confidence score $s\in[0,1]$. They fall into two groups---\emph{single-shot} (one model call) and \emph{multi-signal} (several calls combined)---shown in Fig.~\ref{fig:pipeline}. Throughout, $m$ denotes the number of queries issued per input in the multi-signal group ($m{=}10$).

\subsection{Single-Shot Scores}
\label{sec:single}
\textbf{Verb} is the converted verbalized confidence of Eq.~\eqref{eq:conversion}. When an API exposes token log-probabilities, we additionally form \textbf{Verb-logprob} from the first-token Yes/No log-probabilities (normalized to a class probability), and \textbf{Verb-logprob+T}, the same score after the post-hoc temperature scaling of Section~\ref{sec:calib}. All three cost one model call.

\subsection{Multi-Signal Scores: Building a Feature Vector}
\label{sec:multi}
To obtain richer information, we query the LLM $m{=}10$ times per input and collect one scalar in $[0,1]$ from each query, forming a \emph{feature vector} $\mathbf{x}=(x_1,\dots,x_m)$, where $x_j$ is the score returned by the $j$-th query. We obtain the $m$ queries in two ways.

\paragraph{Interpretable sub-task decomposition.} We hand-design, per dataset, $m{=}10$ named yes/no \emph{sub-questions}, each probing one capability relevant to the decision---e.g.\ \texttt{lexical\_overlap} and \texttt{semantic\_entail} for natural-language inference, or \texttt{explicit\_support} and \texttt{relevance} for question answering (full definitions in the released code). Each sub-question is answered by an independent LLM call that returns a score $x_j\in[0,1]$ \emph{and} a short natural-language rationale that we log. We call these features \emph{interpretable} in an operational sense: each $x_j$ is tied to a human-readable named criterion with a logged rationale, so the score can be explained and audited per example (Section~\ref{sec:interp})---a deployment property, not a ranking claim.

\paragraph{Multi-prompt paraphrase.} As a strong black-box alternative that requires no hand design, we instead generate $m{=}10$ paraphrases of the original classification prompt; each returns a direct classification confidence, converted via Eq.~\eqref{eq:conversion}. These features carry no per-criterion meaning---hence no interpretability---but, as we will see, often rank as well as or better than sub-tasks.

\subsection{Aggregating the Feature Vector}
\label{sec:agg}
Given $\mathbf{x}\in[0,1]^m$ we compare two ways to collapse it to a single score.

\paragraph{Simple averaging.} $\bar{s}=\frac{1}{m}\sum_{j=1}^m x_j$. This implicitly assumes every feature points the same way (higher $\Rightarrow$ more positive) and deserves equal weight---an assumption that fails badly when features have heterogeneous meaning.

\paragraph{Learned combination (logistic regression).} We fit an $L_2$-regularized logistic regression
\begin{equation}
    P(y{=}1\mid\mathbf{x}) = \sigma\!\left(\mathbf{w}^{\top}\mathbf{x}+b\right),
    \label{eq:lr}
\end{equation}
with $\sigma$ the sigmoid and regularization strength $C$ chosen by grid search over five log-spaced values from $0.01$ to $100$. \textbf{The regression is fit on a labeled training split, disjoint from the test set; the exact partition into fitting, calibration, and test sets is given in Section~\ref{sec:setup-exp}} (anticipated here because the split is part of the method, not just the protocol). We name the learned sub-task score \textbf{Sub-LR} and the learned paraphrase score \textbf{MP-LR}. Learned weights matter because features differ in meaning and polarity---on NLI a high \texttt{lexical\_overlap} indicates \emph{entailment}, our negative class---so averaging cancels signal that the regression recovers by giving each feature its correct sign and weight (Section~\ref{sec:probes} quantifies how unreliable the raw probes are alone). The learned constructions thus define both a new score and a new predictor (its $0.5$-threshold decision), which we evaluate end to end.

\paragraph{Diagnostic oracle.} To ask how much signal lives in a few features, we also evaluate, as an \emph{oracle upper bound only}, the best of all $\binom{10}{3}{=}120$ three-feature subsets selected by cross-validated PR-AUC (\textbf{Oracle-3}). It is not a deployable recipe: choosing the wrong three features can be far worse.

\subsection{Calibration}
\label{sec:calib}
The regression logit $z=\mathbf{w}^{\top}\mathbf{x}+b$ may be miscalibrated. We apply temperature scaling $\hat{p}=\sigma(z/T)$, fitting the scalar $T>0$ on a held-out calibration split by minimizing negative log-likelihood~\cite{guo_calibration_2017}. Because $z\mapsto\sigma(z/T)$ is monotone, it cannot change the ranking---PR-AUC is identical before and after---so calibration affects \emph{calibration} and threshold placement, never ranking. We call the calibrated sub-task score \textbf{Sub-cal}.\footnote{We compared temperature with beta calibration~\cite{kull_beta_2017} for the bounded $[0,1]$ inputs: ECE differences were negligible (each better on $4$/$8$ pairs) and PR-AUC identical (both monotone). We keep temperature scaling for simplicity.}

\smallskip
\noindent\textbf{Summary.} The seven constructions we compare are \textbf{Verb}; the naive \textbf{Subtask-avg} and \textbf{MP-avg}; the learned \textbf{Sub-LR}, its calibrated form \textbf{Sub-cal}, and \textbf{MP-LR}; and the diagnostic \textbf{Oracle-3}. For models whose API exposes them we additionally study log-probability scores (\textbf{Verb-logprob}, \textbf{Verb-logprob+T}).

\section{Experiments}
\label{sec:experiments}

\subsection{Setup}
\label{sec:setup-exp}

\paragraph{Datasets.} \textbf{BoolQ}~\cite{clark_boolq_2019} ($1{,}000$ examples, yes/no QA), \textbf{MNLI}~\cite{williams_broad-coverage_2018} ($1{,}000$; non-entailment as the positive class, following the HANS framing~\cite{mccoy_right_2019}), and \textbf{PubMedQA}~\cite{jin_pubmedqa_2019} ($890$; biomedical yes/no), for $n{=}2{,}890$. Positive rates are $62$--$66\%$.

\paragraph{Models.} Nine LLMs of varying strength: \emph{strong} (accuracy ${>}85\%$ on ${\ge}2$ of our benchmarks, cf.\ Table~\ref{tab:appfull})---Llama-3.1-70B, Claude-Haiku-4.5, GPT-4o-mini, GPT-4.1-mini, Gemini-2.5-Flash-Lite---and \emph{weaker}---Gemini-2.5-Flash, Llama-3.1-8B, GPT-4.1-nano, GPT-5-nano. All are queried at temperature $0$. \textbf{Exclusion rule.} A pair enters the study only if the model's responses parse: on BoolQ, Llama-3.1-8B and GPT-4.1-nano produced frequent malformed (unparseable) JSON during data collection and were excluded, leaving $25$ pairs ($7$ BoolQ, $9$ each MNLI/PubMedQA). Models that parse but give near-constant confidence (GPT-5-nano) are \emph{retained}: a constant score is a meaningful granularity outcome ($G_q{=}1$), not a failure.

\paragraph{Data splits and what each metric sees.} For every pair we use a single stratified $70$/$30$ \emph{train}/\emph{test} split. All metrics (PR-AUC, ECE, coverage, AURC, $G_q$, $H_q$) are computed on the held-out \emph{test} portion, identically for every construction, so comparisons are like-for-like. Within the training portion, a further $75$/$25$ split separates \emph{LR fitting} (Eq.~\eqref{eq:lr}, plus its grid search) from \emph{temperature calibration} (Section~\ref{sec:calib}); the test set is never touched during fitting or calibration. Single-shot Verb needs no fitting and is read directly off the test set.

\paragraph{Metrics.} \emph{Class ranking}: PR-AUC (how well $s$ separates the positive from the negative class), and---because a random classifier already scores PR-AUC ${\approx}\pi$ at positive rate $\pi$---the normalized $\mathrm{PR\text{-}AUC}_N=(\mathrm{PR\text{-}AUC}-\pi)/(1-\pi)$ ($0$ random, $1$ perfect). \emph{Calibration}: ECE. \emph{Selective prediction}: $\mathrm{cov}@\tau$ for $\tau\in\{5\%,10\%\}$ and AURC, both computed from the two-sided rule of Section~\ref{sec:deploy} (correctness-based, not class-based). $\mathrm{cov}@\tau$ sweeps a fixed menu of $200$ equally spaced thresholds on $[0,1]$ (step ${\approx}0.005$; the ${\approx}100$ of them lying in $q$'s range $[0.5,1]$ are the effective ones), identically for every construction, reporting the best test-set coverage at risk $\le\tau$---an \emph{oracle} upper bound on \emph{quantized-grid controllability}, not a deployed threshold-selection procedure. Reachable operating points are capped by $\min(G_q,{\approx}100)$ (Section~\ref{sec:resolution}): Verb is score-limited ($G_q\ll100$), aggregated scores are menu-limited ($G_q\gg100$). \emph{Resolution}: $G_q$, $H_q$ (Section~\ref{sec:resolution}). We compute $95\%$ bootstrap CIs ($1{,}000$ resamples) for headline metrics (Verb PR-AUC: mean width $0.04$, range $0.02$--$0.08$); per-cell intervals ship with the released code.

\subsection{Main Result: Good Ranking, Poor Resolution}
\label{sec:main}

\begin{figure}[tbp]
\centering
\includegraphics[width=\textwidth]{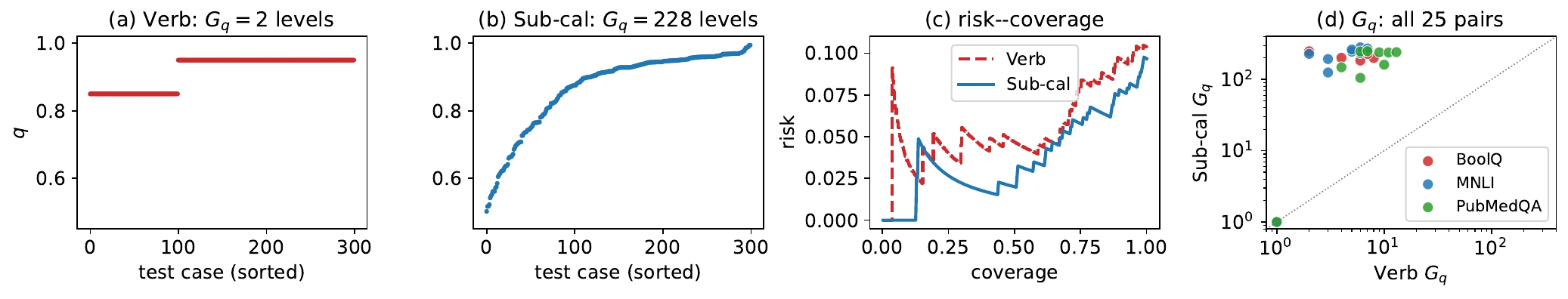}
\caption{\textbf{The score granularity gap} on the thresholded confidence $q$ (MNLI, GPT-4o-mini). (a)~Verb's $4$ class probabilities collapse to only $G_q{=}2$ acceptance levels. (b)~Sub-cal offers $G_q{=}228$ levels at comparable ranking (Table~\ref{tab:appfull}). (c)~Verb's two levels yield a two-step risk--coverage staircase; Sub-cal's curve is near-continuous. (d)~Across all $25$ pairs, Verb clusters at $G_q{=}1$--$13$ while Sub-cal spans $1$--$282$.}
\label{fig:gap}
\end{figure}

The headline finding is that ranking and resolution come apart. Once converted via Eq.~\eqref{eq:conversion}, single-shot \textbf{Verb ranks well}: across the $25$ pairs its $\mathrm{PR\text{-}AUC}_N$ spans $0.41$--$0.95$ (mean $0.80$; full per-pair results in Table~\ref{tab:appfull}). Normalization is essential---raw PR-AUC looks uniformly high ($0.78$--$0.98$) only because the positive rate is high; normalized, strong models ($\mathrm{PR\text{-}AUC}_N{>}0.80$) genuinely rank well while the weakest sit near $0.41$, only moderately above chance.

\begin{table}[tbp]
\centering\scriptsize
\caption{Complete results across all $25$ model--dataset pairs, sorted by accuracy within each dataset. Sub-LR\,/\,MP-LR = sub-task\,/\,multi-prompt features with learned LR; Orc-3 = oracle best-$3$ sub-tasks (diagnostic upper bound). $\Delta=\max(\text{Sub-LR},\text{MP-LR},\text{Orc-3}){-}\text{Verb}$, positive in \textbf{bold}; the two $G_q$ columns give the number of distinct acceptance levels for Verb and Sub-cal.}
\label{tab:appfull}
\setlength{\tabcolsep}{3.5pt}
\renewcommand{\arraystretch}{0.86}
\begin{tabular}{ll ccc cccc c}
\toprule
& & \multicolumn{3}{c}{\textbf{Verbalized (Verb)}} & \multicolumn{4}{c}{\textbf{Multi-signal PR-AUC}} & \textbf{Sub-cal} \\
\cmidrule(lr){3-5}\cmidrule(lr){6-9}\cmidrule(lr){10-10}
\textbf{Dataset} & \textbf{Model} & \textbf{Acc} & \textbf{PR} & \textbf{$G_q$} & \textbf{Sub-LR} & \textbf{MP-LR} & \textbf{Orc-3} & \textbf{$\Delta$} & \textbf{$G_q$} \\
\midrule
\multirow{7}{*}{\rotatebox{90}{\textbf{BoolQ}}}
& GPT-5-nano   & $.623$ & $.812$ & $1$  & $.812$ & $.806$ & $.812$ & $.000$ & $1$ \\
& Gemini-Flash & $.837$ & $.896$ & $2$  & $.748$ & $.857$ & $.817$ & $-.039$ & $248$ \\
& Gemini-Lite  & $.873$ & $.952$ & $4$  & $.909$ & $.958$ & $.948$ & $.006$ & $201$ \\
& GPT-4o-mini  & $.903$ & $.926$ & $7$  & $.956$ & $.971$ & $.964$ & $\mathbf{.045}$ & $228$ \\
& Claude-Haiku & $.920$ & $.974$ & $8$  & $.870$ & $.965$ & $.892$ & $-.010$ & $200$ \\
& Llama-70B    & $.927$ & $.972$ & $6$  & $.934$ & $.976$ & $.960$ & $.004$ & $255$ \\
& GPT-4.1-mini & $.930$ & $.976$ & $6$  & $.772$ & $.961$ & $.818$ & $-.015$ & $184$ \\
\midrule
\multirow{9}{*}{\rotatebox{90}{\textbf{MNLI}}}
& Llama-8B     & $.660$ & $.799$ & $3$  & $.905$ & $.952$ & $.913$ & $\mathbf{.153}$ & $193$ \\
& GPT-5-nano   & $.660$ & $.830$ & $1$  & $.830$ & $.834$ & $.830$ & $.004$ & $1$ \\
& Gemini-Flash & $.723$ & $.851$ & $3$  & $.776$ & $.901$ & $.812$ & $\mathbf{.049}$ & $125$ \\
& GPT-4.1-nano & $.770$ & $.914$ & $7$  & $.931$ & $.956$ & $.946$ & $\mathbf{.042}$ & $271$ \\
& Llama-70B    & $.843$ & $.968$ & $5$  & $.954$ & $.967$ & $.967$ & $-.001$ & $262$ \\
& GPT-4o-mini  & $.897$ & $.963$ & $2$  & $.966$ & $.912$ & $.972$ & $\mathbf{.009}$ & $228$ \\
& GPT-4.1-mini & $.897$ & $.969$ & $6$  & $.913$ & $.969$ & $.930$ & $.000$ & $282$ \\
& Claude-Haiku & $.910$ & $.973$ & $6$  & $.969$ & $.980$ & $.972$ & $.007$ & $234$ \\
& Gemini-Lite  & $.910$ & $.983$ & $5$  & $.967$ & $.983$ & $.984$ & $.001$ & $245$ \\
\midrule
\multirow{9}{*}{\rotatebox{90}{\textbf{PubMedQA}}}
& GPT-5-nano   & $.622$ & $.811$ & $1$  & $.811$ & $.811$ & $.811$ & $.000$ & $1$ \\
& Gemini-Flash & $.640$ & $.776$ & $4$  & $.679$ & $.773$ & $.754$ & $-.004$ & $148$ \\
& GPT-4.1-nano & $.820$ & $.920$ & $7$  & $.830$ & $.918$ & $.867$ & $-.002$ & $249$ \\
& GPT-4o-mini  & $.828$ & $.936$ & $11$  & $.918$ & $.924$ & $.946$ & $\mathbf{.010}$ & $238$ \\
& Gemini-Lite  & $.846$ & $.944$ & $6$  & $.928$ & $.940$ & $.931$ & $-.004$ & $247$ \\
& Llama-8B     & $.854$ & $.943$ & $10$  & $.841$ & $.790$ & $.847$ & $-.096$ & $161$ \\
& Claude-Haiku & $.891$ & $.959$ & $9$  & $.897$ & $.953$ & $.915$ & $-.005$ & $240$ \\
& Llama-70B    & $.903$ & $.948$ & $13$  & $.944$ & $.937$ & $.951$ & $.003$ & $241$ \\
& GPT-4.1-mini & $.906$ & $.953$ & $6$  & $.667$ & $.960$ & $.785$ & $.007$ & $105$ \\
\bottomrule
\end{tabular}
\end{table}

Yet \textbf{Verb is far too coarse}. Measured on the thresholded quantity $q$, a median of $58\%$ of test cases share Verb's single most common acceptance level, and Verb offers only $1$--$13$ distinct levels (median $6$)---fewer than its distinct class probabilities, since $s$ and $1-s$ collapse to the same $q$. By the operating-point bound of Section~\ref{sec:resolution}, this caps an operator at roughly six thresholds---no matter how well the score ranks. Figure~\ref{fig:gap} shows the consequence on one pair: Verb's two acceptance levels yield a two-step staircase (panel c), whereas the learned, calibrated sub-task score spreads $q$ across the range and yields a near-continuous curve.

\paragraph{Resolution is about distribution, not just count.} Counting distinct levels alone could be misleading, so we also report the spread; all summary statistics below and in Table~\ref{tab:summary} are medians across pairs. Verb's acceptance-level entropy is only $H_q{=}1.1$ nats (out of a possible ${\approx}3.0$): the few levels it offers are also unevenly used. Learned, calibrated aggregation moves both quantities together: $G_q$ rises to $228$ (range $1$--$282$), $H_q$ to $2.4$ nats, and the largest tie shrinks from $58\%$ of cases to $3\%$. \textbf{For the sub-task construction this expansion is consistent: aggregation never reduces resolution, and increases it in $22$ of $25$ pairs} (the three exceptions are ties at $G_q{=}1$)---the benefit is resolution, not, as we show next, ranking.

\begin{table}[tbp]
\centering\small
\caption{Summary across $25$ pairs. Aggregation's consistent effect is resolution expansion ($G_q$, $H_q$), not PR-AUC improvement, which is selective. ``PR wins''/``Cov wins'' count pairs where a method's PR-AUC / $\mathrm{cov}@5\%$ exceeds Verb. ``Coll.''\ is the fraction of cases sharing the most common acceptance level. Resolution entries are medians across pairs and are reported for the sub-task variant only.}
\label{tab:summary}
\setlength{\tabcolsep}{5pt}
\begin{tabular}{lcccccc}
\toprule
\textbf{Construction} & \textbf{PR wins} & \textbf{$\Delta$PR} & \textbf{Cov wins} & \textbf{$G_q$} & \textbf{$H_q$ (nats)} & \textbf{Coll.} \\
\midrule
Verb (single-shot)      & ---   & ---       & ---   & $6$   & $1.1$ & $58\%$ \\
Sub-LR / Sub-cal        & $4/25$ & $-.03$   & $11/25$ & $228$ & $2.4$ & $3\%$ \\
MP-LR (paraphrase + LR) & $9/25$ & $.00$    & ---   & ---   & --- & --- \\
\bottomrule
\end{tabular}
\end{table}

\paragraph{Reading token probabilities is a cheap middle ground.} For the two models that expose log-probabilities (Llama-70B, GPT-4o-mini), raw \textbf{Verb-logprob} improves ranking slightly ($+0.003$--$0.050$ PR-AUC), but $65$--$94\%$ of its mass saturates at $q{>}0.99$, where four-decimal values collapse together ($G_q{=}24$--$91$). Post-hoc temperature scaling (\textbf{Verb-logprob+T}, fitted $T{=}1.9$--$6.8$) is strictly monotone in $q$: at exact thresholds it leaves the ranking, the ties, and the entire risk--coverage curve unchanged. All of its benefit is therefore \emph{quantized-grid controllability} (Section~\ref{sec:setup-exp}): dividing logits by $T{>}1$ moves the saturated mass into the region where four-decimal values and the fixed threshold menu can separate it---$G_q$ rises to $46$--$91$, and for the saturated GPT-4o-mini the quantized $\mathrm{cov}@5\%$ jumps from $0$--$5\%$ to $63$--$75\%$, at one model call. The effect is stable across quantizations (at every precision from two to six decimals the calibrated $G_q$ is at least the raw one, e.g.\ $10{\to}30$ at two decimals), but a deployment that stores exact float thresholds would see no coverage gain---only better-calibrated probabilities and a better-spread display scale. The catch is access: $7$ of our $9$ models, and most commercial APIs, do not expose log-probabilities.

\subsection{When Does Multi-Query Aggregation Help---and When Does It Hurt?}
\label{sec:whenhelp}

\begin{figure}[tbp]
\centering
\includegraphics[width=0.78\textwidth]{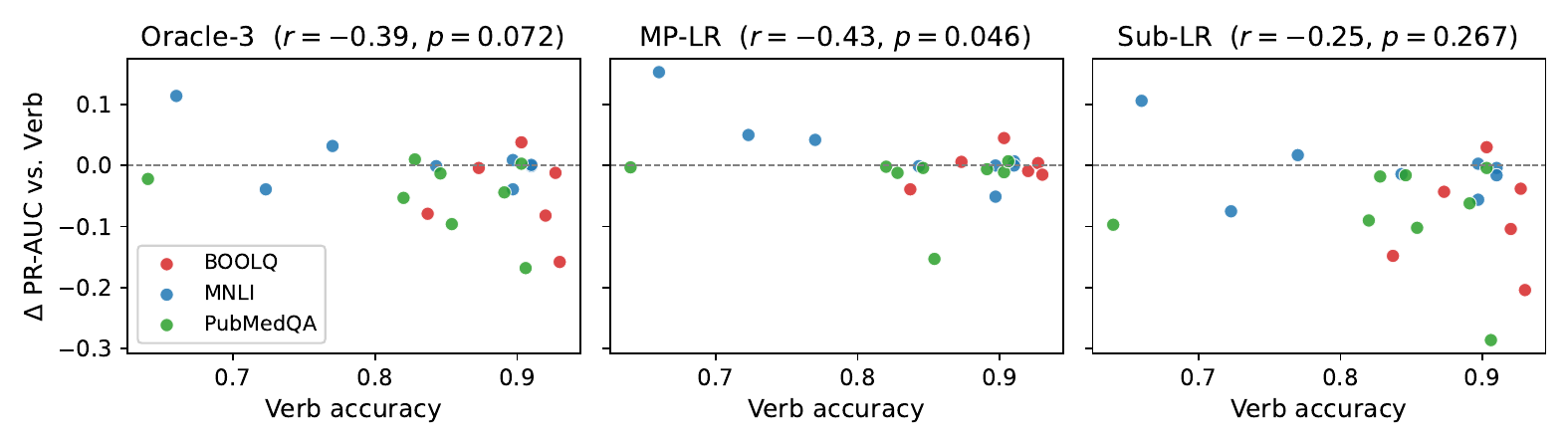}
\caption{Verb accuracy vs.\ PR-AUC gain from aggregation ($n{=}22$ pairs; $3$ GPT-5-nano pairs excluded, $G(\text{Verb}){=}1$). Gains are negatively correlated with model strength---aggregation helps weak models and can hurt strong ones. Colors: BoolQ (red), MNLI (blue), PubMedQA (green).}
\label{fig:delta}
\end{figure}

\begin{figure}[tbp]
\centering
\includegraphics[width=0.74\textwidth]{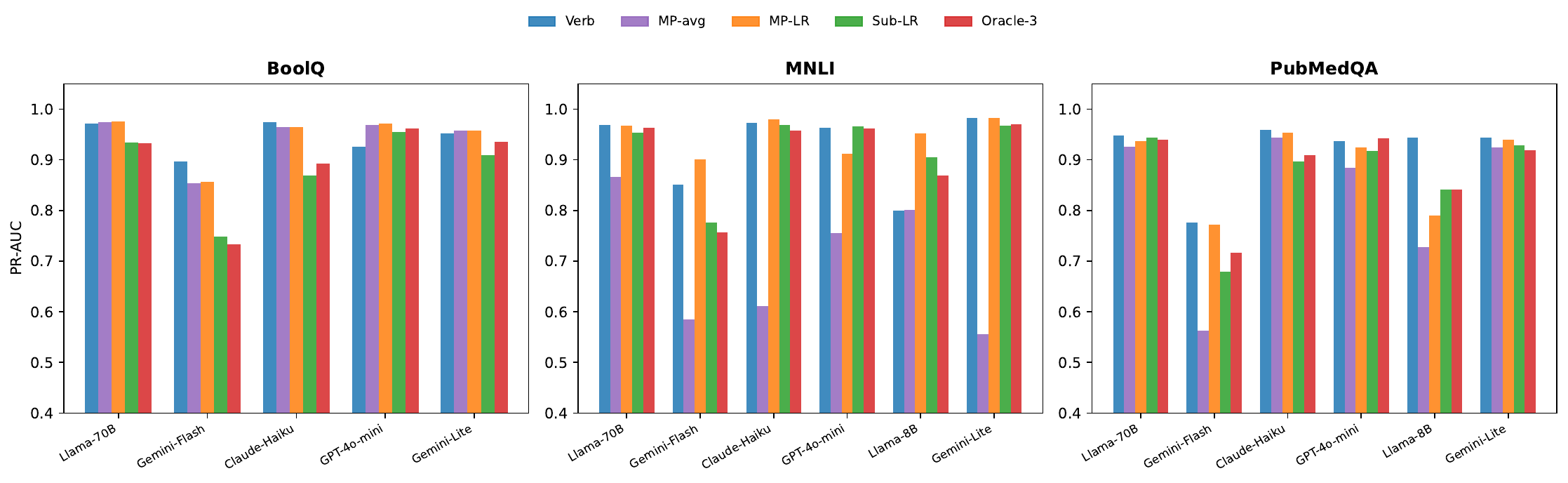}
\caption{PR-AUC of every confidence construction across the $25$ model--dataset pairs. The learned-aggregation methods (Sub-LR, MP-LR) separate from single-shot verbalized confidence mainly in the weaker-model cluster; in the stronger-model cluster all constructions are already near-saturated, so aggregation's practical value there is finer threshold control, not higher ranking.}
\label{fig:allprauc}
\end{figure}

Aggregation's effect on \emph{ranking} is selective, not universal, and---importantly---it is not always positive. Across pairs, Sub-LR beats Verb's PR-AUC in only $4/25$ pairs and MP-LR in $9/25$ (Table~\ref{tab:summary}). The gains concentrate on \emph{weak} models: the correlation between a model's Verb accuracy and the PR-AUC gain from MP-LR is $r{=}{-}0.43$ ($p{=}0.046$, $n{=}22$; Fig.~\ref{fig:delta}). We treat these correlations as exploratory, since the $22$ pairs share models and datasets. Below ${\approx}85\%$ accuracy aggregation gives real ranking gains (e.g.\ Llama-8B/MNLI $+0.153$); above ${\approx}90\%$ the mean change is $-0.048$. Figure~\ref{fig:allprauc} shows this across all constructions: aggregation pulls away from verbalized confidence almost only in the weaker-model cluster.

\paragraph{Aggregation can degrade strong models---we state this plainly.} The deployment consequence is sharper than the ranking averages suggest. Smoothing the score does \emph{not} guarantee a better risk--coverage curve: across all $25$ pairs, AURC favors Verb in $12$ and Sub-cal in $10$ (ties elsewhere), and the degradation \emph{grows with model strength} ($r{=}0.44$, $p{=}0.026$, bootstrap $95\%$ CI $[0.21,0.64]$; exploratory, as above). A strong model's verbalized scores, though coarse, already place high confidence on its correct predictions; injecting learned weights can perturb that good ranking. Sub-cal's AURC advantage appears mainly for genuinely weak models (e.g.\ Llama-8B/MNLI $0.257{\to}0.099$). \textbf{The honest summary is therefore narrower than ``aggregation improves confidence'':} aggregation reliably buys \emph{more operating points} (finer control), and it improves \emph{ranking} only for weak models, while it can \emph{worsen} the risk--coverage of strong ones. We separate three benefits that are easily conflated---(i)~ranking (PR-AUC), (ii)~global risk--coverage (AURC), and (iii)~threshold controllability (number of operating points)---and find that only (iii) improves consistently.

\paragraph{Where decomposition fails outright.} Beyond the strong-model trend, sub-task decomposition has clear individual failures that we do not hide. For GPT-4.1-mini it \emph{collapses}---on BoolQ ($.976{\to}.772$) and PubMedQA ($.953{\to}.667$)---yet the paraphrase variant stays robust on those very pairs ($.961$ and $.960$), showing the failure is specific to the hand-designed probe set, not to multi-query aggregation in general. By contrast, on PubMedQA/Llama-8B \emph{both} variants fall below Verb (Sub-LR $.841$, MP-LR $.790$ vs.\ $.943$): here aggregation itself hurts, consistent with the strong-model degradation trend above. We revisit the interpretable variant's failure modes in Section~\ref{sec:interp}.

\subsection{Ablation: Why Learned Combination Is Necessary}
\label{sec:ablation}

\paragraph{Averaging collapses; learning recovers.} Table~\ref{tab:ablation} isolates the combiner (average vs.\ LR) and the feature type (sub-task vs.\ paraphrase). On MNLI, simple sub-task averaging falls to near chance (PR-AUC $.462$) because the probes have conflicting polarities that cancel; learned combination recovers to $.912$. Even \emph{manually} aligning sub-task polarities before averaging recovers only partly (MNLI $.51{\to}.82$, still below Verb's $.95$): the regression contributes not just sign correction but non-uniform weighting. The pattern is milder on BoolQ and PubMedQA, where polarities are more aligned, but LR dominates throughout.

\begin{table}[tbp]
\centering\small
\caption{Effect of feature type and combiner (PR-AUC, averaged over models). Averaging collapses on MNLI due to polarity conflicts; learned combination recovers it.}
\label{tab:ablation}
\begin{tabular}{lcccc}
\toprule
 & \multicolumn{2}{c}{\textbf{Sub-task}} & \multicolumn{2}{c}{\textbf{Paraphrase}} \\
\cmidrule(lr){2-3}\cmidrule(lr){4-5}
\textbf{Dataset} & \textbf{Avg} & \textbf{LR} & \textbf{Avg} & \textbf{LR} \\
\midrule
BoolQ    & $.844$ & $.857$ & $.895$ & $.928$ \\
MNLI     & $.462$ & $.912$ & $.587$ & $.945$ \\
PubMedQA & $.813$ & $.835$ & $.847$ & $.890$ \\
\bottomrule
\end{tabular}
\end{table}

\paragraph{The individual probes are weak---by design that is fine.} \label{sec:probes}A systematic look at the sub-tasks explains both the averaging collapse and why learning is essential. Treated as standalone classifiers (threshold their own score at $0.5$), the ten MNLI probes are \emph{poor and polarity-confounded}: their naive accuracy against the gold label ranges from $0.26$ (\texttt{semantic\_entail}) to $0.58$ (\texttt{negation\_flip}), most \emph{below} chance---because, e.g., a high entailment score indicates the \emph{negative} class here, so reading it naively is worse than guessing. The learned aggregator turns these confounded signals into a strong score precisely by assigning \texttt{semantic\_entail} and \texttt{lexical\_overlap} large \emph{negative} weights. This is the mechanism behind Table~\ref{tab:ablation}, and it is why no averaging scheme, however polarity-aligned, matches the regression.

\paragraph{Asking for more decimals does not add real resolution.} A tempting shortcut to higher granularity is to prompt the model for finer-grained numbers. It backfires: requesting three-decimal confidences \emph{degrades} PR-AUC in all six tested pairs (e.g.\ PubMedQA/GPT-4o-mini $.936{\to}.580$; MNLI/Llama-70B $.968{\to}.822$). The extra digits are noise, not uncertainty: genuine resolution must come from diverse information, not from formatting.

\paragraph{Robustness.} Three checks confirm the resolution is meaningful, not cosmetic. \emph{Noise is not enough}: jittering Verb scores leaves coverage unchanged in $17$ of the $18$ pairs on which this control was run (subset details in the released code), so the gain is semantic. \emph{Calibration improves too}: Sub-cal has lower ECE than Verb in $15/25$ pairs (e.g.\ Llama-8B/MNLI $.174{\to}.094$). \emph{Label efficiency}: $25\%$ of the training split gives near-full performance; the calibration set ($\ge 40$ examples) is the binding constraint.

\subsection{Interpretability as a Deployment Property}
\label{sec:interp}

Since paraphrase aggregation (MP-LR) ranks as well as or better than sub-task aggregation (Sub-LR) on most pairs (Section~\ref{sec:whenhelp}), the case for the interpretable variant is \emph{not} better ranking. It is the deployment property of Section~\ref{sec:deploy}: \textbf{at a modest ranking cost (Table~\ref{tab:ablation}), interpretable sub-tasks supply transparency the black-box variant cannot}---(i)~\emph{weight transparency} (weights attach to named criteria with sensible signs, and $5$--$7$ of $10$ signs stay stable across $10$ resampled splits); (ii)~\emph{error attribution} (which capability drove a mistake); and (iii)~an \emph{inspectable log} (each call records a rationale a reviewer can read; we do not claim rationales are faithful explanations).

\begin{table}[tbp]
\centering\small
\caption{Two real MNLI examples (GPT-4o-mini) under the model's \emph{shared} learned weights. \textbf{Ex.~A} (\#486, gold: \emph{entailment}, the negative class) and \textbf{Ex.~B} (\#24, gold: \emph{non-entailment}). Verb assigns $\ppos{=}0.85$ to \emph{both}; the decomposition pushes both down (A\,$\to$\,$.088$, B\,$\to$\,$.13$). For A this \emph{corrects} a Verb error; for B it is a confident \emph{mistake}---the same negative weights, opposite correctness, because the entailment probes fire alike but the labels differ.}
\label{tab:case}
\begin{tabular}{lccc}
\toprule
\textbf{Sub-task} & \textbf{LR weight} & \textbf{Ex.~A} & \textbf{Ex.~B} \\
\midrule
\texttt{semantic\_entail}    & $-2.71$ & $0.90$ & $0.90$ \\
\texttt{lexical\_overlap}    & $-2.60$ & $0.75$ & $0.50$ \\
\texttt{negation\_flip}      & $-2.29$ & $1.00$ & $0.90$ \\
\texttt{subsequence}         & $-1.21$ & $0.80$ & $0.90$ \\
\texttt{quantifier\_mismatch}& $+1.63$ & $0.90$ & $0.90$ \\
\bottomrule
\end{tabular}
\end{table}

\paragraph{Error analysis: a success and a failure.} Table~\ref{tab:case} contrasts two examples under identical weights. In Ex.~A the decomposition corrects a confident Verb error; in Ex.~B---where the model's entailment probes fire just as strongly but the gold label is \emph{non}-entailment---the very same negative weights drive a confident \emph{mistake}. This is decomposition's characteristic failure: it trusts probe \emph{patterns} that do not always track the label, and the individually unreliable probes (Section~\ref{sec:probes}) leave nothing to veto the error. With the pair-level failures of Section~\ref{sec:whenhelp} (\emph{which} models collapse) and the per-probe analysis (\emph{which} signals are weak), this is a systematic account of when and why decomposition errs. The interpretable representation does not prevent these failures, but it makes them \emph{visible}---a contradicted probe, an unstable weight---whereas a paraphrase ensemble fails silently.

\section{Deployment Recommendations}
\label{sec:reco}

No construction dominates; each occupies a distinct point in the trade-off of ranking, resolution, inference cost, and interpretability. We distill the evidence into a decision procedure (cost in calls per input: Verb and logprob $1{\times}$, Oracle-3 $3{\times}$, multi-query aggregation $10{\times}$).

\begin{enumerate}
\item \textbf{Always} convert verbalized confidence to a class probability (Eq.~\eqref{eq:conversion})---free, and it removes a systematic distortion.
\item \textbf{If log-probabilities are available}, use them with post-hoc temperature scaling: at $1{\times}$ cost this avoided the worst staircase effects in our data.
\item \textbf{If the model is weak} (here, ${\lesssim}85\%$ accuracy), multi-query aggregation often improves ranking and consistently expands resolution, at $10{\times}$ cost.
\item \textbf{If the model is strong} (${\gtrsim}90\%$), use Verb for risk--coverage and add aggregation \emph{only} when fine threshold control is specifically needed, since aggregation can worsen AURC here (cut-offs are descriptive of our data).
\item \textbf{If interpretability is required}, prefer the sub-task variant: for a modest ranking cost relative to MP-LR it adds weight transparency, error attribution, and inspectable logs.
\end{enumerate}

\paragraph{Cost, and how to evaluate.} Aggregation's $10{\times}$ calls are a real hurdle at scale, and the gap does not close by scaling the model---even frontier-class models offer only $2$--$13$ distinct acceptance levels---so calibrated log-probabilities are the better value where exposed. Finally, our results argue for reporting \emph{two} axes, not one---ranking (PR-AUC, AURC) \emph{and} resolution ($G_q$, $H_q$), with the normalized $\mathrm{PR\text{-}AUC}_N$: a score that ranks perfectly but offers six thresholds is not deployment-ready, and reporting ranking alone hides this.

\section{Limitations}
\label{sec:limits}

We study \emph{binary} classification on three English benchmarks with $62$--$66\%$ positive rates; the granularity gap under severe class imbalance, multi-class, or generative settings, and in other languages~\cite{xue_mlingconf_2025}, remains open. Our study is comparative, not exhaustive: semantic-entropy families~\cite{kuhn_semantic_2023}, conformal wrappers~\cite{angelopoulos_conformal_2025}, and deep ensembles are not benchmarked. The interpretable variant needs per-domain manual sub-task design and labeled data (${\ge}155$ examples); MP-LR avoids the manual design but forfeits interpretability; both cost $10{\times}$ inference. Finally, calibrated log-probabilities---the best cost/resolution trade-off we found---could be evaluated on only $2$ of $9$ models, narrowing that conclusion.

\section{Conclusion}
\label{sec:conclusion}

A confidence score can rank cases almost perfectly and still be unfit for deployment if it offers an operator only a handful of thresholds. We made this \emph{score granularity gap} precise---bounding operating points by the number of distinct acceptance levels of the thresholded confidence $q$ and measuring their spread with entropy---and used it to compare seven confidence constructions across $25$ model--dataset pairs. Single-shot verbalized confidence ranks surprisingly well but is far too coarse; token log-probabilities with calibration widen the gap cheaply where available; and multi-query aggregation widens it the most but at $10{\times}$ cost, helping weak models while sometimes degrading strong ones. Interpretability, we argued, is best seen as a deployment property, earned by sub-task scores at a modest ranking cost relative to paraphrase aggregation. We hope future evaluations report resolution alongside ranking, so confidence scores are judged by whether they can actually be operated.

\smallskip
\noindent All code, cached predictions, and sub-task definitions will be released upon publication.

\begin{credits}
\subsubsection{\ackname}
We used Claude (Anthropic) as a writing assistant during manuscript preparation; all technical content, experiments, and conclusions are the authors' own.
\subsubsection{\discintname}
The authors have no competing interests to declare.
\end{credits}

\smallskip
\noindent\textbf{Reproducibility.} Complete per-pair results for all $25$ model--dataset pairs are in Table~\ref{tab:appfull}. Sub-task definitions, the high-precision-prompting experiment, additional diagrams, and implementation details (structured JSON output at temperature $0$; $L_2$ logistic regression with $C$ grid-searched over five log-spaced values; scalar temperature via NLL minimization) accompany the released code. All code and cached predictions will be released upon publication.

\bibliographystyle{splncs04}
\bibliography{llm_cls_zotero}

\ifdefined\ARXIV
\appendix
\section{Additional Diagrams}
\label{app:figs}

\begin{figure}[H]
\centering
\includegraphics[width=0.49\textwidth]{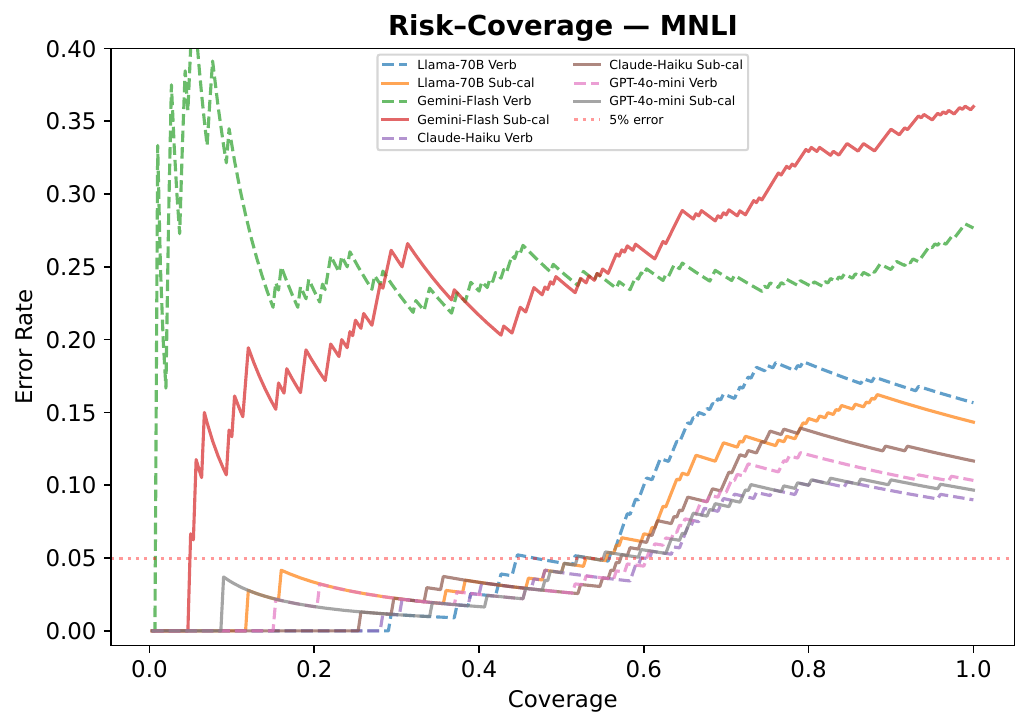}\hfill
\includegraphics[width=0.49\textwidth]{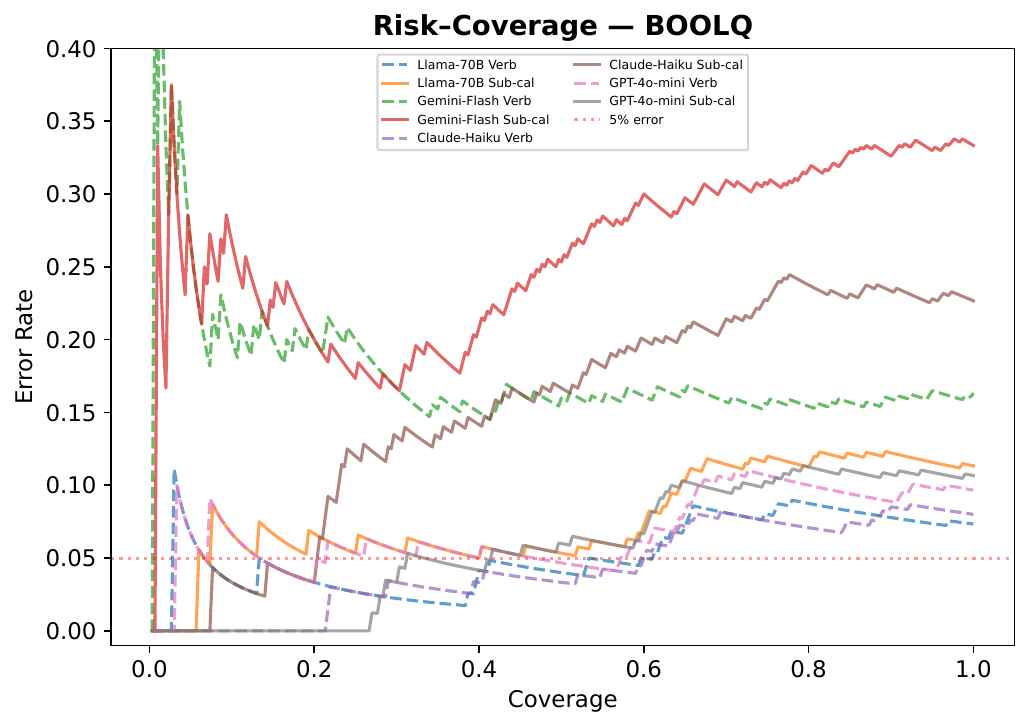}
\caption{Risk--coverage curves for MNLI (left) and BoolQ (right). Verbalized confidence (dashed, staircase-shaped) vs.\ calibrated sub-task aggregation (solid, smooth).}
\end{figure}

\begin{figure}[H]
\centering
\includegraphics[width=0.49\textwidth]{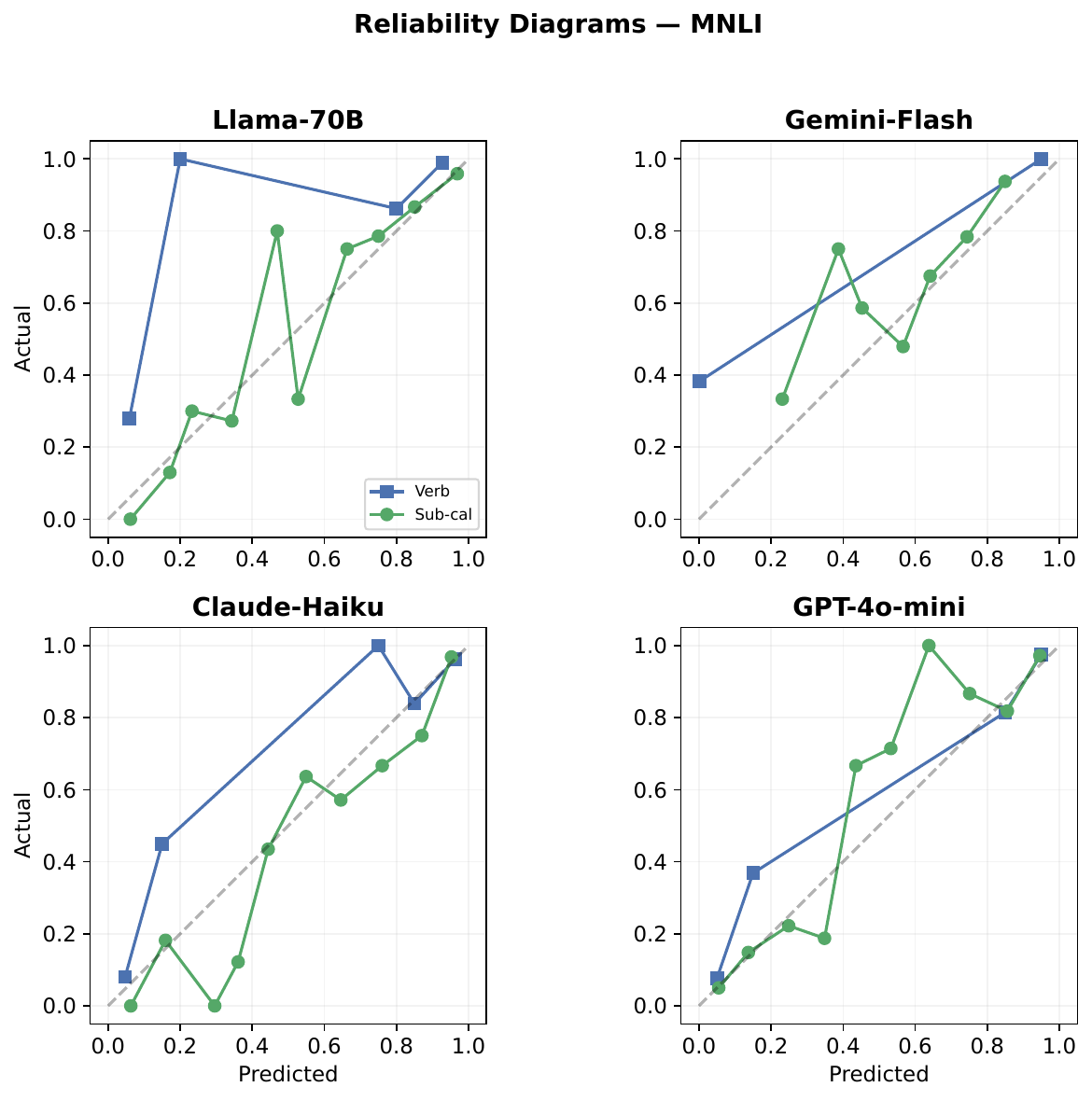}\hfill
\includegraphics[width=0.49\textwidth]{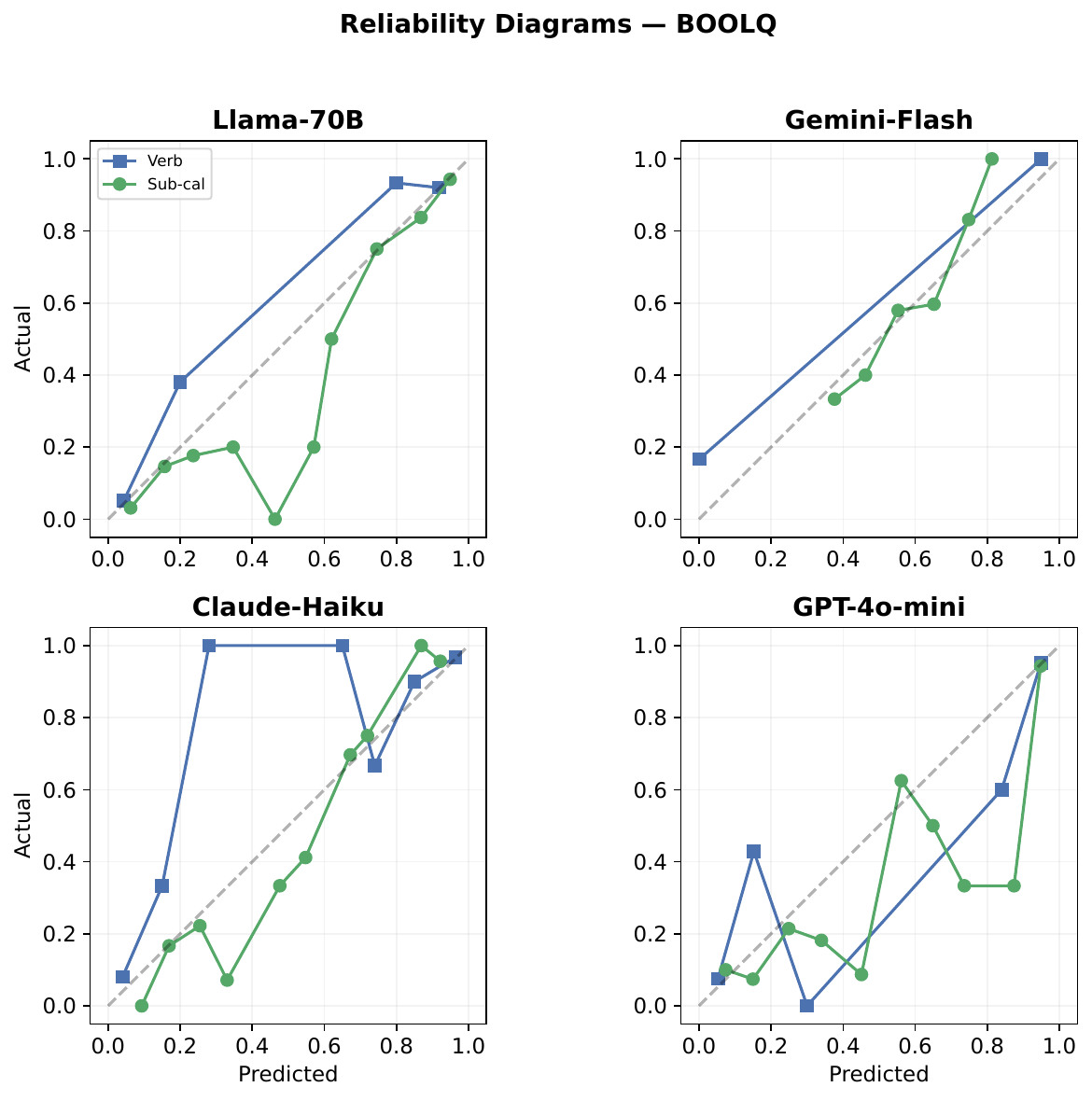}
\caption{Reliability diagrams for MNLI (left) and BoolQ (right). Calibrated sub-task aggregation generally tracks the diagonal more closely than verbalized confidence.}
\end{figure}

\begin{figure}[H]
\centering
\includegraphics[width=0.7\textwidth]{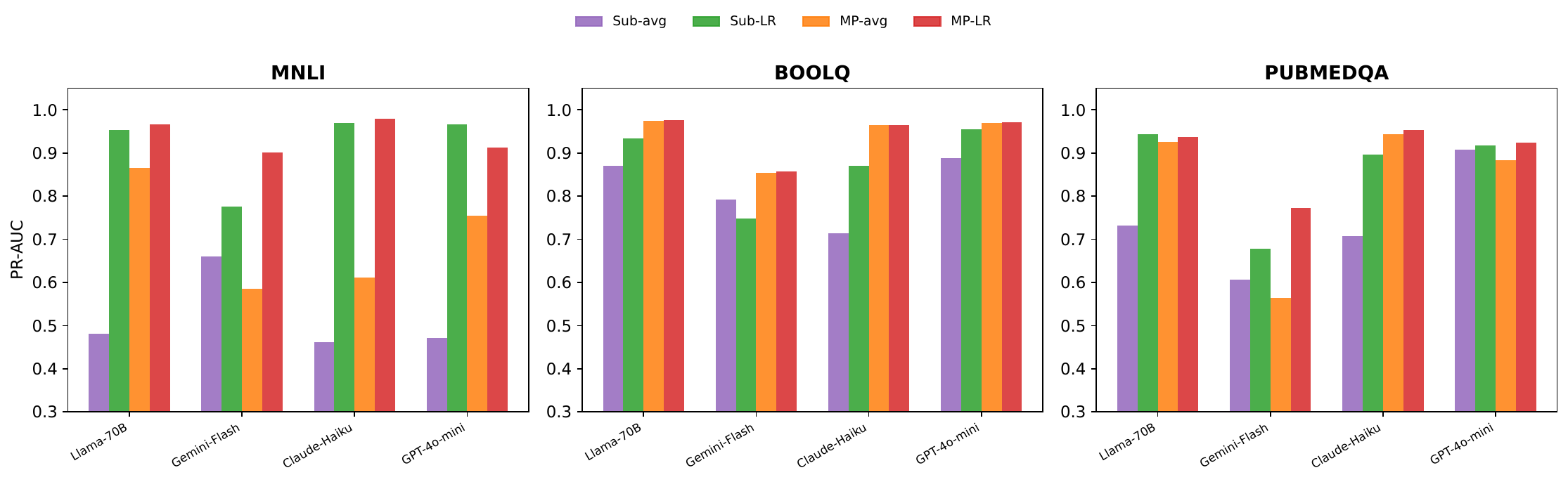}
\caption{Aggregation comparison (average vs.\ logistic regression $\times$ sub-task vs.\ multi-prompt). On MNLI, simple averaging collapses to near chance due to polarity conflicts among sub-task signals.}
\end{figure}

\begin{figure}[H]
\centering
\includegraphics[width=0.49\textwidth]{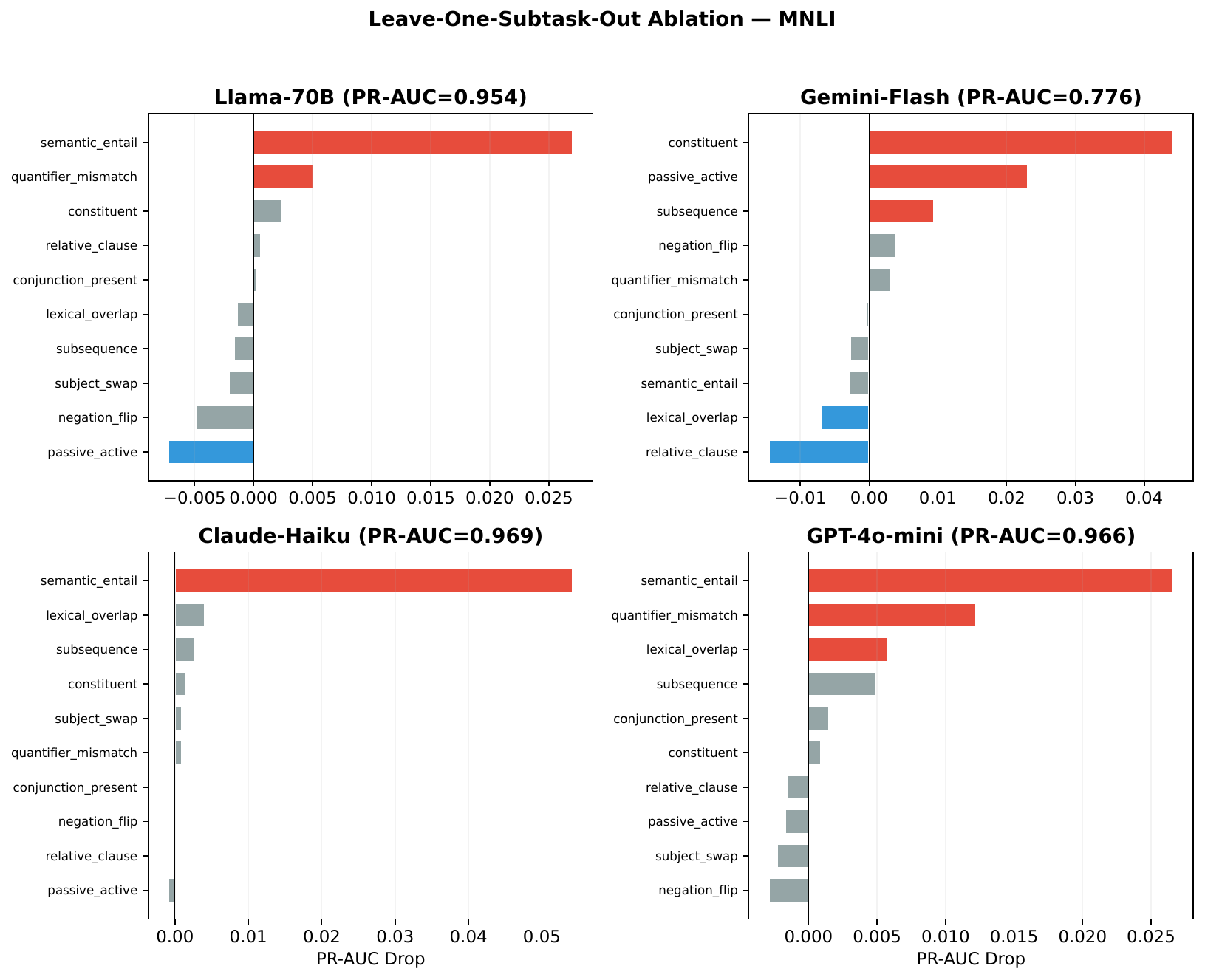}\hfill
\includegraphics[width=0.49\textwidth]{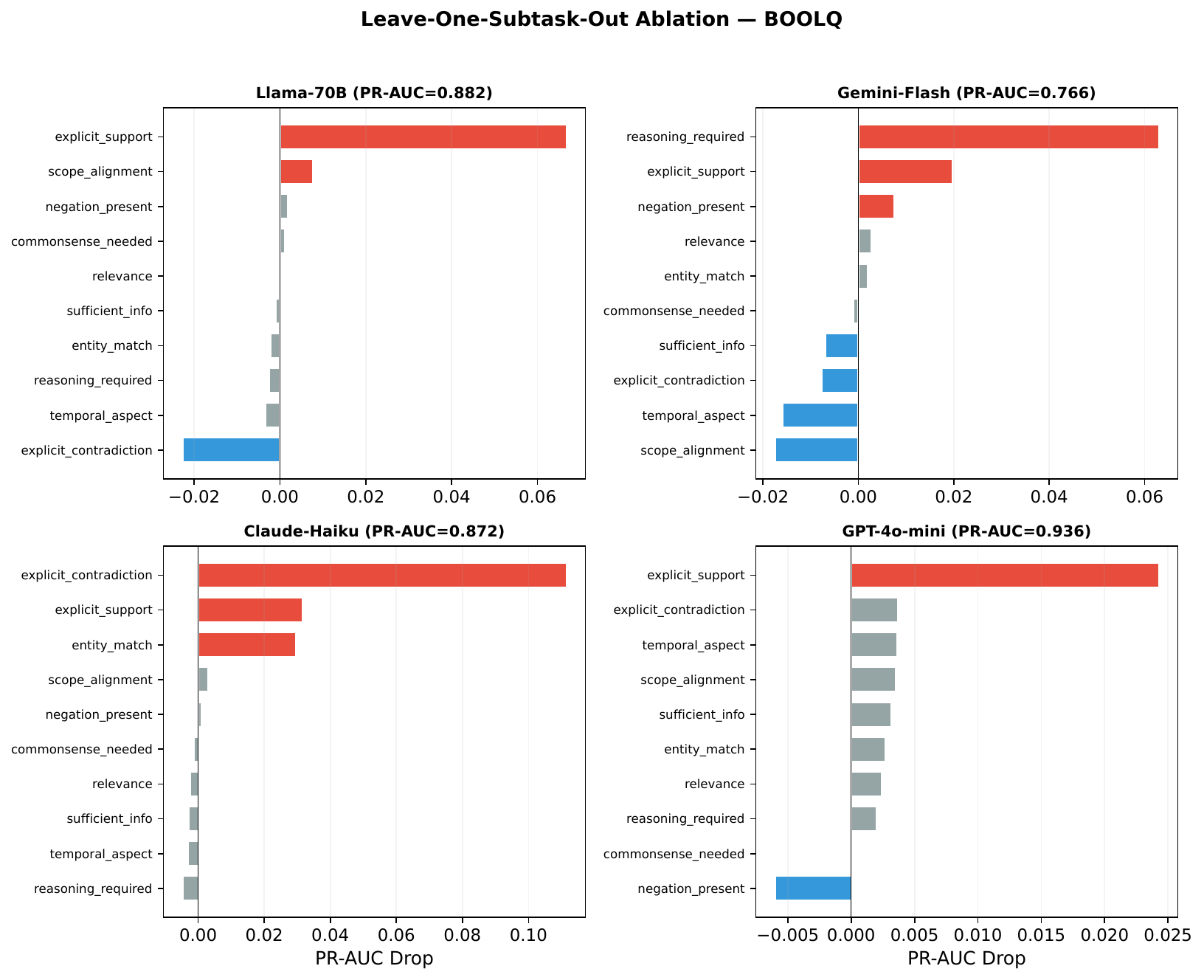}
\caption{Leave-one-out ablation on MNLI (left) and BoolQ (right): \texttt{semantic\_entail} is the most impactful single feature for most strong models on MNLI.}
\end{figure}

\begin{figure}[H]
\centering
\includegraphics[width=0.32\textwidth]{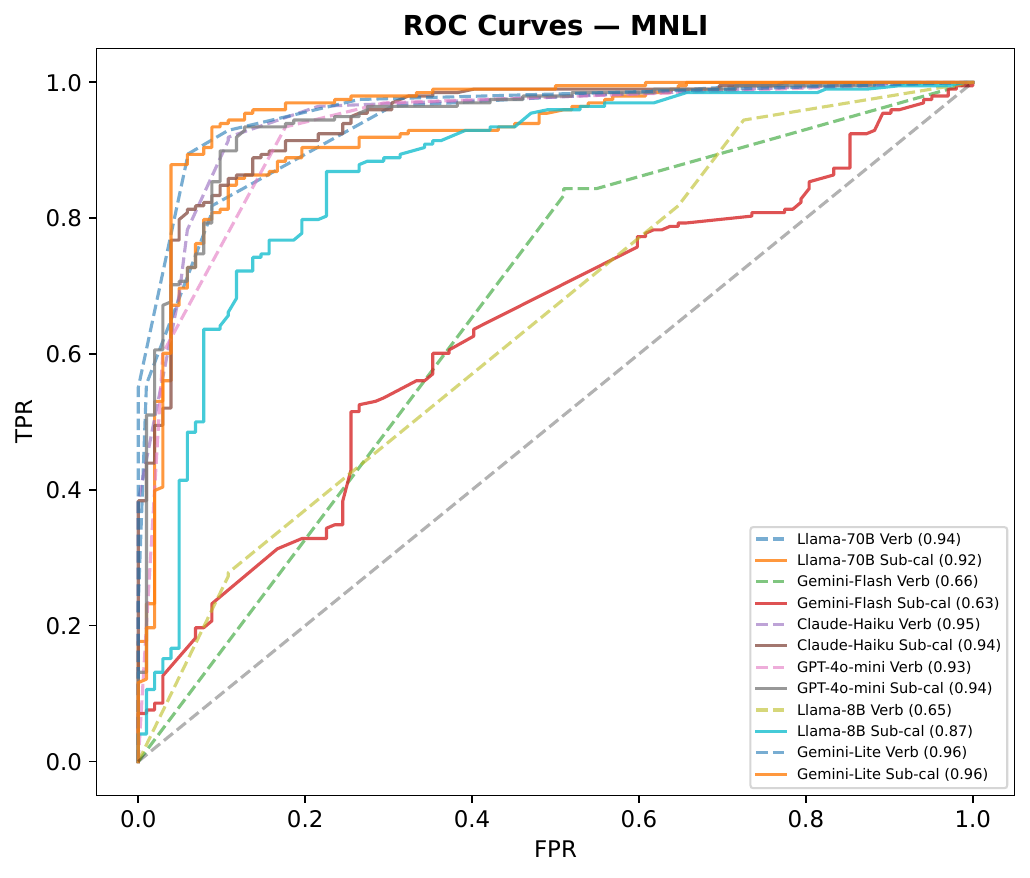}\hfill
\includegraphics[width=0.32\textwidth]{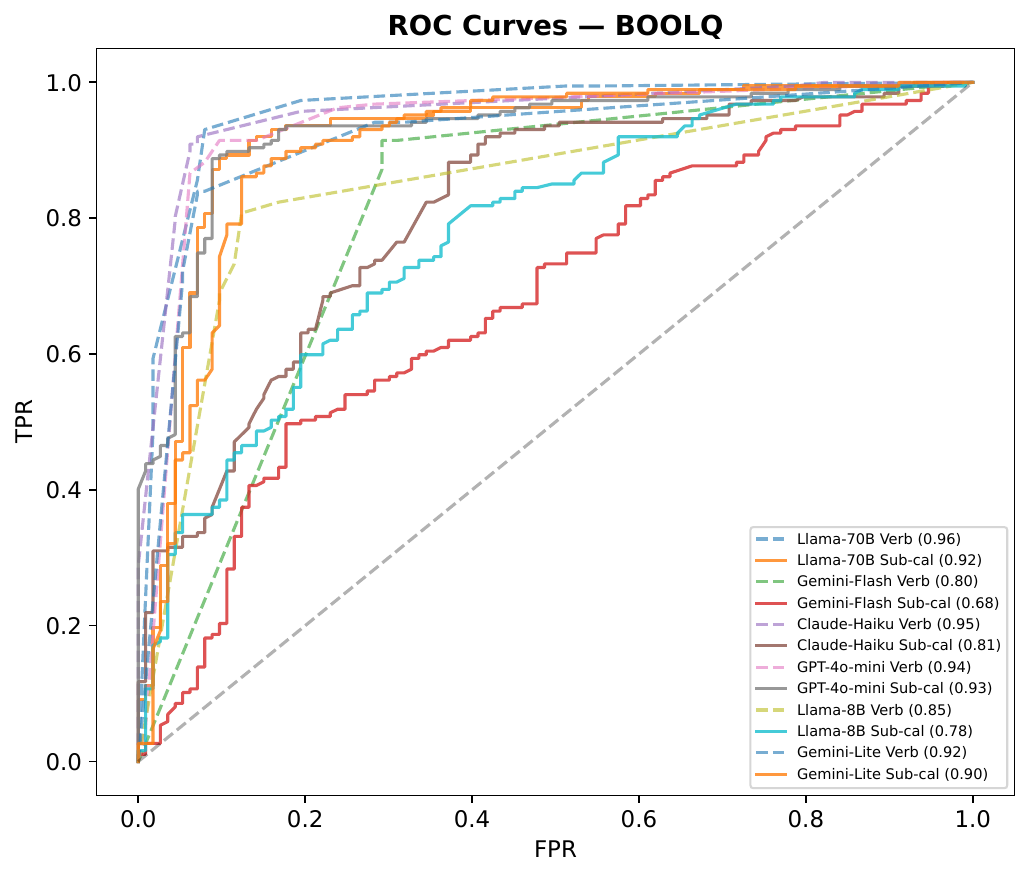}\hfill
\includegraphics[width=0.32\textwidth]{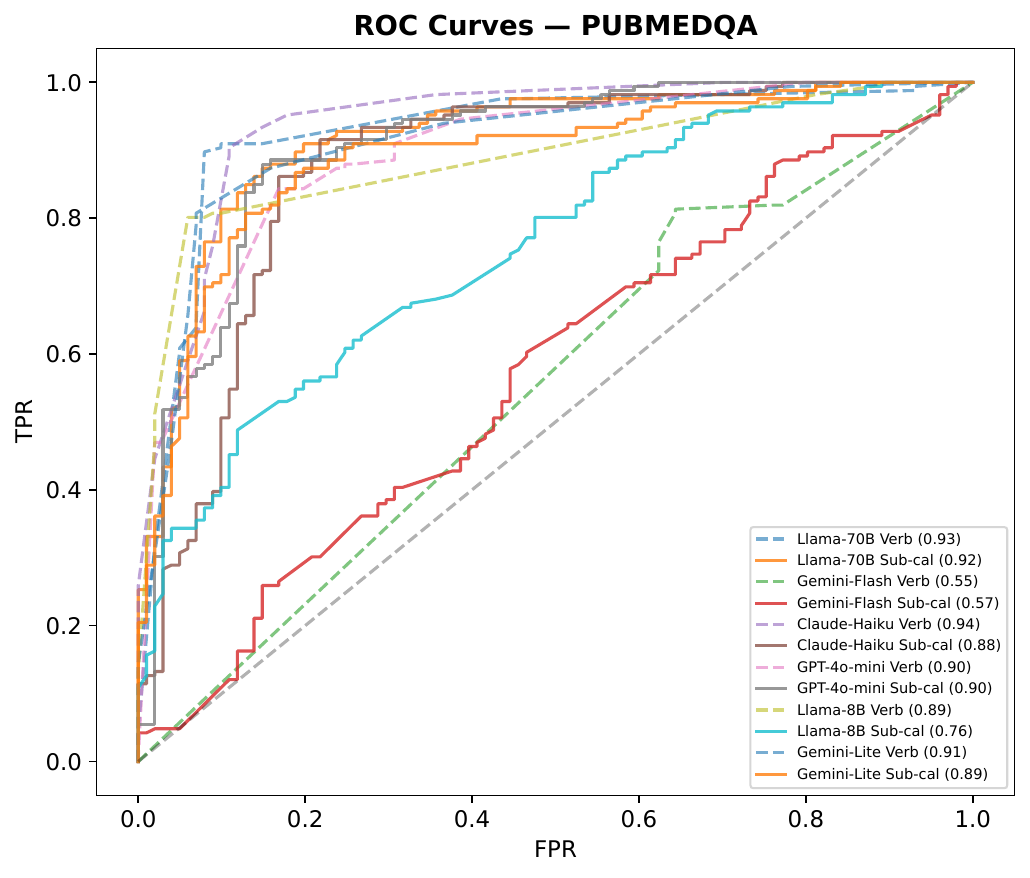}
\caption{ROC curves for MNLI, BoolQ, and PubMedQA across confidence constructions.}
\end{figure}
\fi

\end{document}